\documentclass[runningheads]{llncs}

\usepackage{eccv}

\usepackage{eccvabbrv}

\usepackage{graphicx}
\usepackage{booktabs}

\usepackage[accsupp]{axessibility}  

\usepackage{hyperref}

\usepackage{orcidlink}

\usepackage{booktabs}
\usepackage{array}
\usepackage{indentfirst} 
\usepackage{algorithm}
\usepackage{algpseudocode} 
 \usepackage{multirow}
 \usepackage{colortbl}
\usepackage{tcolorbox}
\usepackage{enumitem}
\usepackage{wrapfig}
\usepackage{mdframed}
\usepackage{tabularx}
\usepackage{booktabs}
\usepackage{multirow}

\makeatletter
\renewcommand{\@fnsymbol}[1]{%
  \ensuremath{\ifcase#1\or
  \star\or\dagger\or\ddagger\or
  \S\or\P\or\|\or
  \star\star\or\dagger\dagger\or\ddagger\ddagger
  \else\@ctrerr\fi}}
\makeatother

\newcommand{\corrmark}{\textsuperscript{\ensuremath{\dagger}}}

\begin{document}

\title{SIFT: Self-Imagination Fine-Tuning for Physically Plausible Motion in Video Diffusion Models} 

\titlerunning{SIFT}

\author{
Ruoyu Wang\inst{1} \and
Jialun Liu\inst{2}\thanks{Project leader}\thanks{Corresponding authors} \and
Huayang Huang\inst{1} \and
Haibin Huang\inst{2} \and 
Jiepeng Wang\inst{2} \and
Chi Zhang\inst{2} \and
Xuelong Li\inst{2} \and
Yu Wu\inst{3}\corrmark
}
\authorrunning{R.~Wang et al.}

\institute{
School of Computer Science, Wuhan University
\\
 \and
Institute of Artificial Intelligence, China Telecom (TeleAI)
%
\\
 \and
School of Artificial Intelligence, Wuhan University
}

\maketitle

\begin{abstract}
Recent advances in video diffusion models have greatly improved visual fidelity, yet their generated motions often violate physical plausibility. We observe a common kinematic failure, ``motion entanglement'', the unintended coupling of independent motion sources, such as camera movement and object motion. 
We identify that this issue stems from data bias and the reconstruction-based training design of diffusion models.
Training on noisy videos that still retain coarse motion cues inadvertently encourages the model to replicate existing motion without an incentive to learn how to model kinematically-grounded motions.
To address this, we propose a Self-Imagination Fine-Tuning (SIFT) paradigm, which enables the model to learn from its own generated videos rather than directly reconstructing real ones, breaking the reconstruction shortcut.
We further employ motion-aware discriminative supervision and a progressive hard-case replay strategy to stabilize and accelerate learning.
By leveraging freely-generated text prompts, our method can densely cover a broad motion space, including rare or finely-disentangled scenarios that would be costly to collect as video data. 
Extensive experiments demonstrate that our approach substantially improves the physical realism, motion disentanglement, and controllability of generated videos.
\end{abstract}

\section{Introduction}
\label{sec:intro}

\begin{figure}[t]
    \centering
     \includegraphics[width=\textwidth]{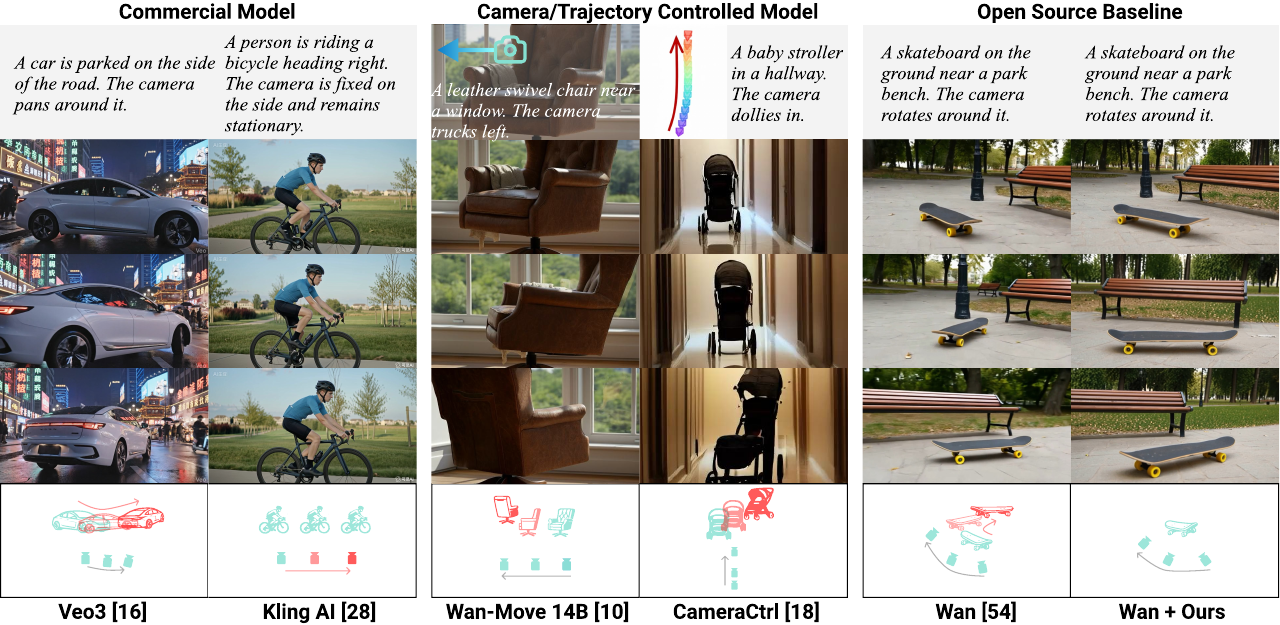}
     \caption{Illustration of Motion Entanglement. The first row shows the input conditions, and the last row visualizes the generated trajectories where red indicates physically implausible relative motion. Note that camera-control methods take the camera trajectory and the first frame as extra conditions, while the others are purely text-to-video. }
    \label{fig: teaser}
\end{figure}



While Video Diffusion Models (VDMs)~\cite{hong2022cogvideo,wan2025wan,yang2024cogvideox,ma2025stepvideo,kong2024hunyuanvideo,liu2024sora,zheng2024opensora} have achieved remarkable visual fidelity and semantic consistency, the physical plausibility of their generated motions remains a fundamental challenge.
Existing efforts~\cite{tan2024physmotion,kang2024howfar,liu2024physgen,bansal2024videophy} largely center on \textit{dynamics}, such as gravity, collisions, and fluid mechanics. In this work, we study a complementary yet highly visible failure from a \textit{kinematics} perspective: \textit{Motion Entanglement}.
This phenomenon refers to the model's inability to independently control and disentangle motions originating from distinct sources, such as camera movement versus object motion. Fundamentally, this reflects a failure to preserve independent reference frames and accurately model relative motion.
As illustrated in \cref{fig: teaser}, this issue is pervasive across commercial models (Veo3~\cite{veo3_2025}, Kling AI~\cite{klingai_2025}), open-source models (Wan~\cite{wan2025wan}), and even models with explicit camera (CameraCtrl~\cite{he2024cameractrl}) or trajectory (Wan-Move~\cite{chu2025wan-move}) controls.
For example, when the camera is supposed to orbit around a stationary object, the generated video frequently shows the object drifting as well. Conversely, if the object moves but the camera should remain fixed, the camera often unintentionally tracks the object.
Although individual frames look realistic, their kinematically flawed relative trajectories reveal a profound lack of structural motion understanding in existing VDMs. Beyond perceptual realism, such kinematic entanglement severely undermines the utility of VDMs as world simulators for downstream applications like autonomous driving, where strict decoupling of ego-motion and environmental dynamics is essential.

\textbf{Why does motion entanglement arise?}  
We attribute this phenomenon to two main factors.
First, \textit{a pronounced data-induced bias exists}: real-world videos are largely composed of scenes where camera and object motions co-occur, and large-scale datasets~\cite{chen2024panda,ju2024miradata,nan2024openvid,wang2025koala,wang2023internvid,bain2021frozen-webvid} lack explicit annotations that distinguish their contributions. Consequently, models learn spurious statistical correlations, treating these independent kinematic variables as inherently linked.
Second, \textit{the training paradigm of VDMs inherently limits learning of kinematic reasoning}. 
VDMs are trained to reconstruct a clean video from a noisy counterpart using pixel-level objectives like MSE loss.
However, even a heavily noised video still preserves substantial residual structural and temporal cues~\cite{wang2025noisequery,wu2024freeinit,hou2024trainingcamtrol,mao2023guidedinitnoise}.
This creates a ``reconstruction shortcut'': the model learns to replicate preexisting motion patterns inherited from the noisy input rather than inferring kinematically correct dynamics from the textual prompt from scratch.
This renders the models inherently insensitive to motion semantics.
Meanwhile, the pixel-level reconstruction objective further biases learning toward appearance fidelity. Each pixel’s temporal displacement can stem from both camera-induced scene movement and object-intrinsic motion, but the loss only enforces accurate RGB matching. Thus, the model learns to replicate the overall visual distribution rather than disentangling the underlying kinematic reference frames.

\textbf{Why traditional fine-tuning fails?} 
A seemingly straightforward solution would be to perform supervised fine-tune (SFT) on additional motion-decoupled video data. 
However, such data is extremely scarce, and curating a dataset to cover the full diversity of real-world motion scenarios would require prohibitive data engineering efforts.
More fundamentally, SFT inherits the same flaws as pretraining: the denoising shortcut and pixel-level reconstruction bias remain intact.
This demonstrates that the motion is being \textit{inherited} from the input data, not \textit{inferred} from the semantics of the prompt, rendering standard SFT ineffective for teaching true motion reasoning.

To overcome these limitations, we introduce a novel paradigm shift from reconstruction-based training to imagination-based learning. Our Self-Imagination Fine-Tuning (SIFT) framework aims to enable the model to learn to understand, correct, and model motion from its own generated videos, rather than merely copying motion patterns from real videos that are inherently physically correct.
To achieve this, we discard real video inputs and force the model to generate videos from randomly initialized noise, guided solely by textual prompts. This completely removes the input reconstruction shortcut, compelling the model to derive kinematic relationships strictly from semantic intent.
Additionally, we replace the pixel-level reconstruction objective with motion-aware discriminative supervision to refine its motion generation toward kinematically consistent dynamics. To further stabilize training and improve learning efficiency, we introduce a progressive hard-case replay strategy that gradually exposes the model to increasingly difficult self-imagined samples. 
Crucially, the text prompts can be freely generated by large language models to densely cover diverse and even rare motion scenarios without requiring paired or motion-decoupled video data.
Experiments demonstrate that our method significantly improves the physical plausibility and disentanglement of generated motion.

In summary, our contributions are as follows:
\begin{itemize}
\item We identify and analyze the Motion Entanglement in video diffusion models, revealing how their current training pipeline limits learning kinematically-grounded motion.
\item We propose a novel self-imagination training paradigm that removes the reconstruction shortcut and enables learning of physically consistent, disentangled motion from textual descriptions.
\item We conduct extensive experiments demonstrating that our framework significantly improves motion realism and relative dynamics, without requiring motion-decoupled video-text data.
\end{itemize}

\section{Related Work}
\label{sec:related_work}

\noindent \textbf{Text-to-Video Generation.}
Text-to-video (T2V) generation has undergone rapid development in recent years, with diffusion models emerging as the predominant backbone.
Early video diffusion models~\cite{blattmann2023stablevideodiffusion,guo2023animatediff,chen2023videocrafter1,chen2024videocrafter2,khachatryan2023text2video,ho2022imagenvideo,hou2024trainingcamtrol} typically ``inflate'' 2D generative models into the video domain by directly adding temporal attention layers to the U-Net architecture of pre-trained text-to-image models (T2I)~\cite{rombach2022high-ldm,ramesh2022hierarchical-dalle,saharia2022photorealistic-imagen,nichol2021glide} to achieve inter-frame coherence. Despite their simplicity and efficiency, these methods often suffer from limited temporal modeling capacity, leading to issues like frame flicker, motion jitter, or inconsistent object evolution over time.
A major shift came with the advent of Diffusion Transformers (DiTs)~\cite{peebles2023scalabledit}, which replaced the U-Net’s backbone with a full transformer architecture to enable 3D full attention. Benefiting from this architectural innovation and advances in large-scale training, recent DiT-based T2V models~\cite{ma2024latte,wan2025wan,hong2022cogvideo,yang2024cogvideox,kong2024hunyuanvideo,liu2024sora,ma2025stepvideo,zheng2024opensora} have achieved remarkable breakthroughs in visual-fidelity, generating high-resolution videos with detailed textures, accurate semantic alignment to text prompts, and improved temporal smoothness. 
Nevertheless, current T2V methods often struggle with physically consistent motion dynamics. This limitation highlights that visual fidelity and semantic alignment alone are insufficient for generating truly realistic and controllable videos; a deeper, physically-grounded motion understanding is required.

\noindent \textbf{Physics-Grounded Video Generation.}
Although video generation models have made significant progress in visual quality, a fundamental challenge is becoming increasingly prominent: the generated dynamic content often lacks physical rationality~\cite{kang2024howfar,bansal2024videophy,meng2024towardsworld-phy,zhang2025videorepa}.
The research community is trying to solve this problem from multiple directions, though most existing efforts primarily focus on \textit{dynamics}.
(1) Simulation-driven generation~\cite{liu2024physgen,zhang2024physdreamer,tan2024physmotion,xie2025physanimator} integrates explicit physical simulation (rigid-body, elastodynamics, material modeling) with generative video rendering. For instance, PhysGen~\cite{liu2024physgen} simulates the rigid-body motion and interactions of each instance based on Newton’s Laws and physical constraints.
(2) LLM-guided or semantic-planning approaches~\cite{lv2024gpt4motion,lian2023llm-phy,xue2025phyt2v,yang2025towards-phy-vlm,zhang2025thinkdiffphy} leverage large language models or motion planning modules to script object trajectories or physical interactions, which are then used to condition the video generation model.
While both directions mark important progress toward physical realism, they typically inject external physical knowledge or simulate dynamics beforehand, treating the generative video model merely as a renderer that translates pre-computed trajectories into pixels.
Furthermore, they largely overlook the fundamental \textit{kinematic} failures inherent in the models' underlying reference frames.
In contrast, our work shifts the focus to kinematics, endowing VDMs with internalized physical reasoning to resolve motion entanglement. We train models to imagine and synthesize motion directly from semantic intent, without relying on external physics engines or pre-simulated trajectories.

\section{Method}
\label{sec:method}
\subsection{Preliminaries}


Text-to-video (T2V) generation aims to synthesize high-quality videos that are semantically aligned with user prompts. Most state-of-the-art approaches build on diffusion models, which generate videos by iteratively denoising a latent representation conditioned on a text prompt $p$. During training, noise $\epsilon$ is added to the real video $x_0$ sampled from the training dataset to produce a noisy version $x_t$. For instance, in standard discrete diffusion models (e.g., DDPMs~\cite{ho2020denoising-ddpm,song2020denoising-ddim,song2020score}), this process is formulated as:
\begin{equation}
x_t = \sqrt{\alpha_t} \, x_0 + \sqrt{1-\alpha_t} \, \epsilon, \quad \epsilon \sim \mathcal{N}(0, I),
\label{eq:forward_ddpm}
\end{equation}
where $\alpha_t$ is cumulative noise schedule. Alternatively, Flow Matching-based models~\cite{liu2022flow,geng2025meanflow,lipman2022flow} instead define a linear interpolation between the data and noise:
\begin{equation}
x_t = (1-t) \, x_0 + t \, \epsilon, \quad t \in [0, 1].
\label{eq:forward_flow}
\end{equation}

The diffusion model $f_\theta$ denoises $x_t$ by predicting $\epsilon$, $x_0$, or a velocity field $v$, but in practice these variants typically share a pixel-level MSE training objective:
\begin{equation}
\mathcal{L}_{\text{MSE}} = \mathbb{E}_{x_0, \epsilon, t} \Big[ \| f_\theta(x_t, t, p) - y_t \|^2 \Big],
\label{eq:mse_loss}
\end{equation}
where $y_t$ denotes the corresponding target (noise, video, vector field). 
As we will demonstrate,  this objective encourages accurate low-level signal reconstruction, but it does not directly enforce reasoning about the underlying motion structure.

\subsection{Motivation} \label{sec: toy}

\begin{figure}[t]
  \centering
  \includegraphics[width=\linewidth]{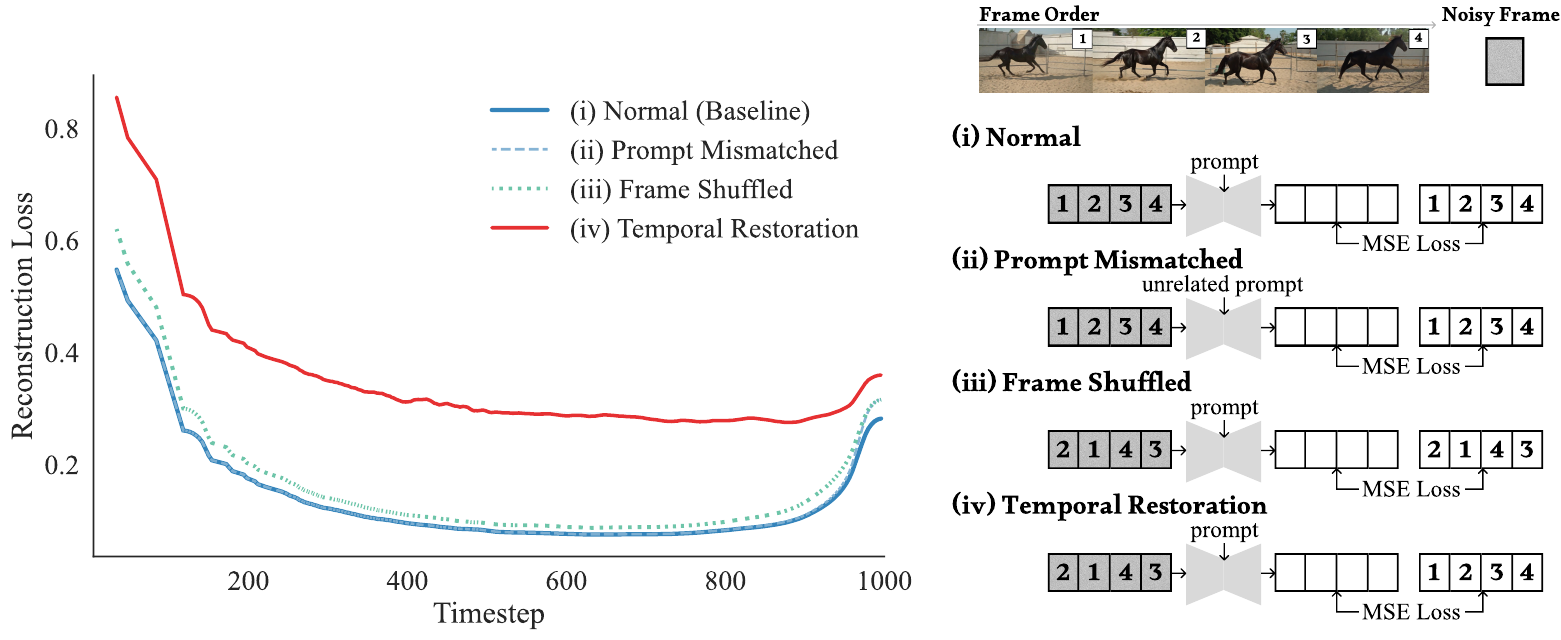}
   \caption{
   MSE loss curves of four input settings. 
   This experiment analyzes pretrained diffusion model behavior in the training setting (one-step prediction). Because the loss is dominated by shortcut cues in the noisy input, it is largely insensitive to prompt and frame-order perturbations. This explains why standard SFT is ineffective for learning physically plausible motion generation and motivating our Self-Imagination paradigm.
   }
   \label{fig: toy}
\end{figure}

Recent studies~\cite{wang2025noisequery,wu2024freeinit,burgert2025go,hou2024trainingcamtrol,mao2023guidedinitnoise} have demonstrated that noisy inputs in diffusion models retain substantial residual information from the original data.
This residual information acts as a ``reconstruction shortcut'' during training, allowing the model to recover the clean video by replicating preexisting motion rather than by reasoning from the conditioning signals (e.g., textual prompts).
This effect is particularly pronounced in pre-trained diffusion models, which already possess strong low-level reconstruction abilities.

To verify this hypothesis, we conduct a simple diagnostic experiment inspired by Videojam~\cite{chefer2025videojam}.
Specifically, we evaluate a pretrained Wan2.1-T2V-1.3B~\cite{wan2025wan} model under four distinct input settings to assess how sensitive its reconstruction behavior is to different conditioning factors:
\begin{itemize}[leftmargin=*,nosep]
\item (i) \textit{Normal}: the original video with its correct prompt;
\item (ii) \textit{Prompt Mismatched}: the original video with an unrelated prompt;
\item (iii) \textit{Frame Shuffled}: a temporally shuffled video with the correct prompt;
\item (iv) \textit{Temporal Restoration}: a shuffled video with the correct prompt, with reconstruction loss computed against the original temporally coherent video.
\end{itemize}

In each case, we add random noise levels $t \in [0,1]$ (corresponding to 0-1000 diffusion timesteps) to 1,000 video-text pairs and compute the reconstruction loss (velocity prediction for Wan) at each timestep. 
As shown in \cref{fig: toy}, the first three settings yield almost identical loss curves, indicating that the model relies predominantly on the residual information in the noisy input rather than the text prompt or meaningful temporal order. 
In contrast, setting (iv) yields significantly higher losses, suggesting that the model has difficulty addressing the temporal dynamics when the prompt strongly contradicts the residual motion cues inherent in the noisy input.

These results reveal that the model behaves largely as a pixel-level refiner rather than a physically grounded generator. Such reliance on reconstruction shortcuts fundamentally limits the effectiveness of conventional supervised fine-tuning (SFT), which may require massive data yet still fails to improve motion understanding.
This motivates us to break the reconstruction shortcut and force the model to learn motion modeling from prompts, which we achieve through our proposed self-imagination fine-tuning.

\subsection{Self-Imagination Fine-tuning} 
\begin{figure}[t]
  \centering
  \includegraphics[width=\linewidth]{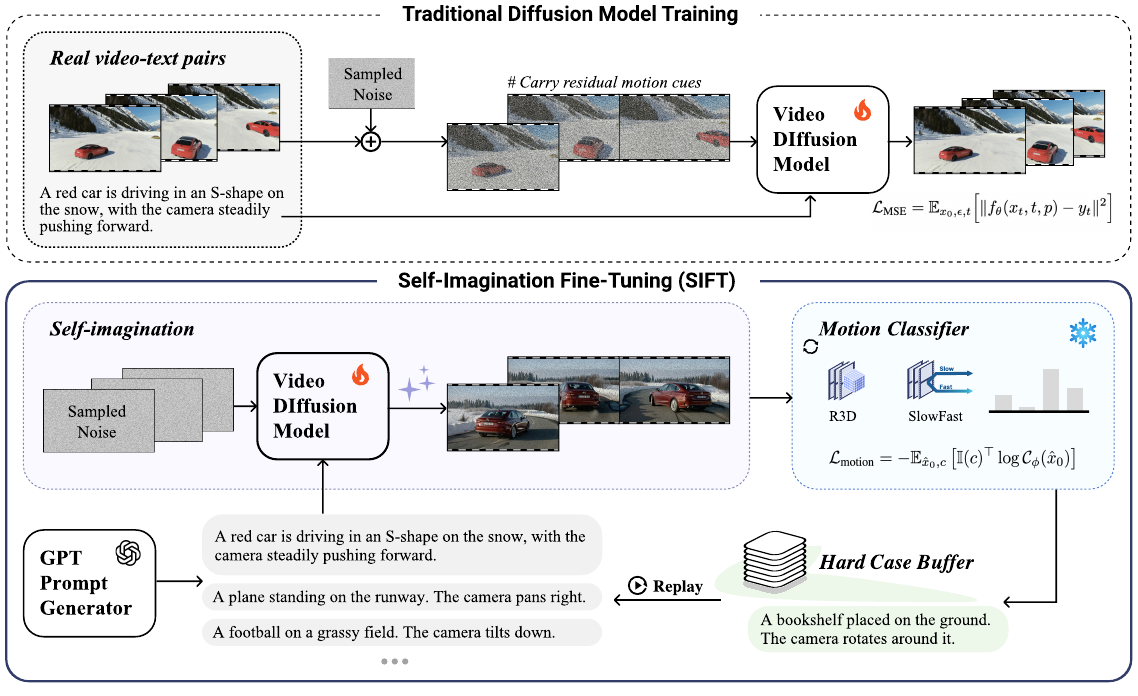}
   \caption{Pipeline comparison between traditional diffusion model training (top) and our proposed self-imagination fine-tuning (bottom).}
   \label{fig:pipeline}
\end{figure}

This section presents an overview of our method. The pipeline is shown in \cref{fig:pipeline}, and pseudocode is provided in \cref{alg:self-imagination}. For clarity, the VAE encoder/decoder and classifier-free guidance (CFG) are omitted.

Compared to standard reconstruction-based training of diffusion models, we discard real video inputs and replace them with pure Gaussian noise. Therefore, the model needs to imagine the video and construct motion solely from textual prompts that specify scene content and motion relations.
These text prompts, which can be freely produced by large language models (e.g., GPT~\cite{hurst2024gpt4o}), provide an unlimited source of training scenarios spanning diverse and even rare combinations of relative motions that are difficult to obtain from real videos. 
To encourage the imagined motions to be physically plausible, we replace the primary reconstruction objective in the self-imagination branch with motion-aware discriminative supervision, while retaining a lightweight MSE term on real video-text pairs to preserve visual quality.
Through this process, the model learns to infer and synthesize coherent, physically plausible motion dynamics directly from semantic intent.

In practice, since diffusion models establish global motion structure primarily during early denoising stages~\cite{wu2025customcrafter,ling2024motionclone}, we restrict training to high-noise regimes and perform only a few denoising steps (e.g., 3 steps) from pure noise.
This allows us to efficiently probe the model's motion imagination where it matters most, significantly reducing computational cost compared to full-sequence generation.

\begin{algorithm}[t]
\caption{Self-Imagination Fine-Tuning}
\label{alg:self-imagination}
\begin{algorithmic}[1]
\footnotesize
\Require Pretrained VDM $f_\theta$, motion classifiers $\mathcal{C}_\phi$ (R3D \& SlowFast), prompt generator, real batch $\mathcal{D}_{\mathrm{real}}$, Hard Case Buffer $\mathcal{B}$
\State Initialize $\mathcal{B} \gets \emptyset$
\For{each training iteration $s$}
    \State $\mathcal{L}_{\mathrm{motion}} \gets 0$
    \State Generate a batch of prompts $\{(p_i, c_i)\}$
    \For{each $(p_i, c_i)$ in the batch}
        \State $x_T \sim \mathcal{N}(0, I)$
        \State $\hat{x}_0 \gets f_\theta^{\mathrm{denoise}}(x_T, p_i)$
        \State $c_{\mathrm{pred}} \gets \mathcal{C}_\phi(\hat{x}_0)$
        \State $\mathcal{L}_{\mathrm{motion}} \gets \mathcal{L}_{\mathrm{motion}} - \mathbb{I}(c_i)^\top \log c_{\mathrm{pred}}$
        \If{$\arg\max(c_{\mathrm{pred}}) \neq c_i$}
            \State Add $(p_i, c_i)$ to $\mathcal{B}$
        \EndIf
    \EndFor
    \State Sample a real batch and compute $\mathcal{L}_{\mathrm{MSE}}$
    \State $\mathcal{L}_{\mathrm{total}} \gets \lambda \mathcal{L}_{\mathrm{motion}} + \mathcal{L}_{\mathrm{MSE}}$
    \State Update $\theta \gets \theta - \eta \nabla_\theta \mathcal{L}_{\mathrm{total}}$
    \State Sample cases from $\mathcal{B}$ with probability $p_s=\min(1, s/S_{\mathrm{warmup}})$ and replay them
\EndFor
\end{algorithmic}
\end{algorithm}

\subsection{Decoupled Motion Feedback}
To provide motion-specific supervision rather than low-level appearance guidance, we train motion classifiers $\mathcal{C}_\phi$ to categorize each video into four motion types: \textit{camera-only motion}, \textit{object-only motion}, \textit{both in motion}, and \textit{both static}.
Since the motion type for each self-generated video is known from its original LLM-generated prompt, we obtain ground-truth labels automatically without manual annotation.
To mitigate model bias and capture complementary inductive priors, we employ two heterogeneous classifiers.
The first is a \textit{3D-ResNet (R3D)}~\cite{tran2018closer-r3d}, which treats spatial and temporal dimensions symmetrically through unified 3D convolutions, making it adept at modeling short-term local motion coherence.
The second is a \textit{SlowFast} network~\cite{feichtenhofer2019slowfast}, which explicitly factorizes temporal and spatial processing into dual pathways operating at different frame rates, one focusing on detailed spatial semantics and the other on high-frequency motion dynamics.
This architectural dichotomy ensures our motion feedback encompasses both specialized and unified temporal understandings.

During training the classifiers, we explicitly simulate the noisy inference of SIFT: adding heavy noise to clean videos ($t\!\in\![900,1000)$) and running the same one-step denoising to obtain predicted $\hat{x}_0$. Classifiers are trained on these noisy, regenerated $\hat{x}_0$, which match the distribution they see when supervising SIFT and ensure robustness to noise and artifacts. On a held-out validation set constructed under the same noisy supervision distribution, R3D and SlowFast achieve 78.4\% and 82.8\% accuracy, respectively.

To further prevent overfitting to a single classifier, we employ an \textit{alternating supervision strategy}. Rather than averaging their outputs, we alternate between the two classifiers across different training batches. This ensures the model benefits from both spatial and temporal inductive biases. The motion loss is then defined using cross-entropy between the predicted and true motion classes:
\begin{equation}
\ell(\hat{x}_i, c_i) = -\, \mathbb{I}(c_i)^\top \log \mathcal{C}_\phi(\hat{x}_i),
\end{equation}
where $\mathbb{I}(c)$ is the one-hot vector of the ground-truth motion type.

\subsection{Progressive Hard Case Replay}
To further improve stability and maximize learning efficiency, we introduce a \textit{progressive hard case replay strategy}. The key idea is to gradually expose the model to increasingly difficult motion cases, rather than overwhelming it with hard samples from the beginning.
After each imagination step, videos that fail to produce correct motion categories are stored in a hard case buffer $\mathcal{B}$.
During the early training phase, these hard samples are temporarily excluded from gradient updates to avoid unstable optimization caused by noisy supervision. 
As training progresses, hard cases are incorporated into the loss with a gradually increasing participation $p_s$:
\begin{equation}
p_s = \min\Big(1, \frac{s}{S_\text{warmup}}\Big),
\end{equation}
where $s$ is the current training iteration step and $S_\text{warmup}$ is a predefined warm-up period.
Formally, let $b_s$ denote the current batch and $H \subseteq \mathcal{B}$ be the set of hard samples identified by motion classifiers. For each sample $i \in b_s$, we draw $u_i \sim \text{Uniform}[0,1]$, and include it in the loss only if $u_i \le p_s$. The batch-wise loss is defined as:
\begin{equation}
\mathcal{L}_\text{motion} = 
 \sum_{i \in b_s}
\Big[
\mathbf{1}_{\{i \notin H\}} 
+ \mathbf{1}_{\{i \in H\}} \mathbf{1}_{\{u_i \le p_s\}}
\Big]
\, \ell(\hat{x}_i, c_i),
\end{equation}
where $\ell(\hat{x}_i, c_i)$ is the motion classification loss for the generated video $\hat{x}_i$ with ground-truth motion label $c_i$.

Meanwhile, we periodically replay samples from the buffer to reinforce challenging motion patterns.
This progressive replay strategy allows the model to first stabilize its understanding on borderline cases before tackling the complex motion dynamics, leading to more robust and stable convergence.

\section{Experiment}
\label{sec:experiment}

\subsection{Experimental Setup}
\noindent\textbf{Dataset.}
To train motion classifiers, we first construct a dataset comprising 4,000 videos, evenly distributed across four motion categories: camera-only motion, object-only motion, both in motion, and both static. 
The videos are manually filtered and collected from existing large-scale video datasets~\cite{xue2025ultravideo,nan2024openvid,lin2025camerabench,fu20243dtrajmaster-360motion,gillman2025force} and open-source websites~\cite{pexels}. Each video is annotated with two types of captions: (1) general content descriptions generated using Qwen-VL-2.5~\cite{bai2025qwen2}, and (2) detailed camera movement descriptions generated by a specialized model~\cite{lin2025camerabench} fine-tuned on Qwen-VL-2.5 with camera-specific data.

\noindent\textbf{Benchmark.}
For testing, we focus on the two critical categories that require disentanglement: camera-only motion and object-only motion.
Our full test set comprises 100 test prompts, each following a precise two-clause format: \{content prompt\} \{camera prompt\}.
The camera clause always begins with ``The camera...'' and encompasses a diverse set of 12 motion templates (e.g., pan, tilt, truck, orbit, etc.) and 5 static templates.
For moving object scenarios, content prompts are sampled from the motion binding subset of T2V-CompBench~\cite{sun2025t2v-compbench}. For static object scenarios, content prompts are generated using Gemini-Pro-Vision-1.5~\cite{team2024gemini}, covering a comprehensive range of entities including vehicles, common objects, and living beings.

\begin{table}[t]
    \centering
    \small
    \renewcommand{\arraystretch}{1.1}  
    \caption{Quantitative comparison of our method with baselines across VLM and human evaluations. Metrics include Semantic Adherence (SA) and Physical Commonsense (PC). Our method achieves the highest scores in all categories.}
    \begin{tabularx}{\textwidth}{l@{\extracolsep{\fill}}cccccccc} 
  \toprule
        \multirow{3}{*}{\textbf{Model Setting}} & \multicolumn{4}{c}{\textbf{VLM Score ($\uparrow$)}} & \multicolumn{4}{c}{\textbf{Human Score ($\uparrow$)}} \\
        & \multicolumn{2}{c}{\textbf{Camera-only}} & \multicolumn{2}{c}{\textbf{Object-only}} & \multicolumn{2}{c}{\textbf{Camera-only}} & \multicolumn{2}{c}{\textbf{Object-only}} \\
        \cmidrule(lr){2-3} \cmidrule(lr){4-5} \cmidrule(lr){6-7} \cmidrule(lr){8-9} 
        & \textbf{SA} & \textbf{PC} & \textbf{SA} & \textbf{PC} & \textbf{SA} & \textbf{PC} & \textbf{SA} & \textbf{PC} \\
        \midrule
        Wan~\cite{wan2025wan} & 3.58 & 4.06 & 3.80 & 3.60 & 3.00 & 2.79 & 2.91 & 2.75 \\
        ~~+ SFT & 3.94 & 4.73 & 4.32 & 4.30 & 3.26 & 3.53 & 2.91 & 2.43 \\
        ~~+ SIFT (Ours) & \textbf{4.80} & \textbf{4.93} & \textbf{4.75} & \textbf{4.72} & \textbf{3.89} & \textbf{4.24} & \textbf{3.98} & \textbf{3.84} \\
        \midrule
        CogVideoX~\cite{yang2024cogvideox} & 3.95 & 3.10 & 3.75 & 2.81 & 3.51 & 3.20 & 2.73 & 2.39 \\
        ~~+ SFT & 4.68 & 4.32 & 4.25 & 3.69 & 3.50 & 3.10 & 3.18 & 2.86 \\
        ~~+ VideoREPA~\cite{zhang2025videorepa} & 4.53 & 4.05 & 4.00 & 3.50 & 3.34 & 4.05 & 2.20 & 2.60 \\
        ~~+ SIFT (Ours) & \textbf{4.89} & \textbf{4.84} & \textbf{4.38} & \textbf{4.25} & \textbf{3.88} & \textbf{4.25} & \textbf{3.75} & \textbf{3.68} \\
        \bottomrule
    \end{tabularx}
    \label{tab: main_results}
\end{table}
\begin{figure}[!t]
  \centering
  \includegraphics[width=0.58\textwidth]{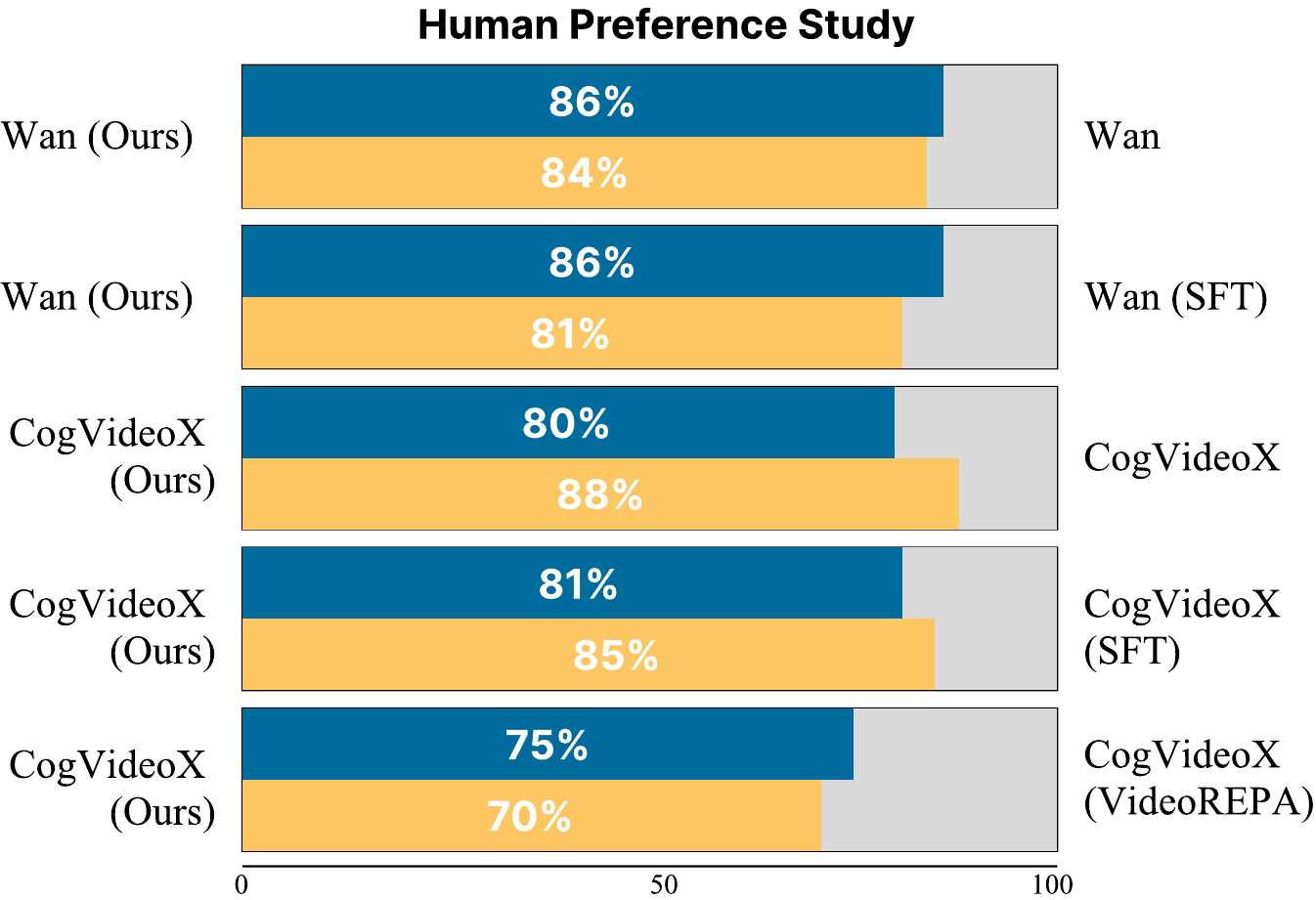}
   \caption{Human preference study results, showing the winning rate of our method vs. baselines. Blue bars indicate Semantic Adherence (SA) and yellow bars indicate Physical Commonsense (PC).}
   \label{fig: preference}
\end{figure}

\begin{figure}[t]
  \centering
    \includegraphics[width=\linewidth]{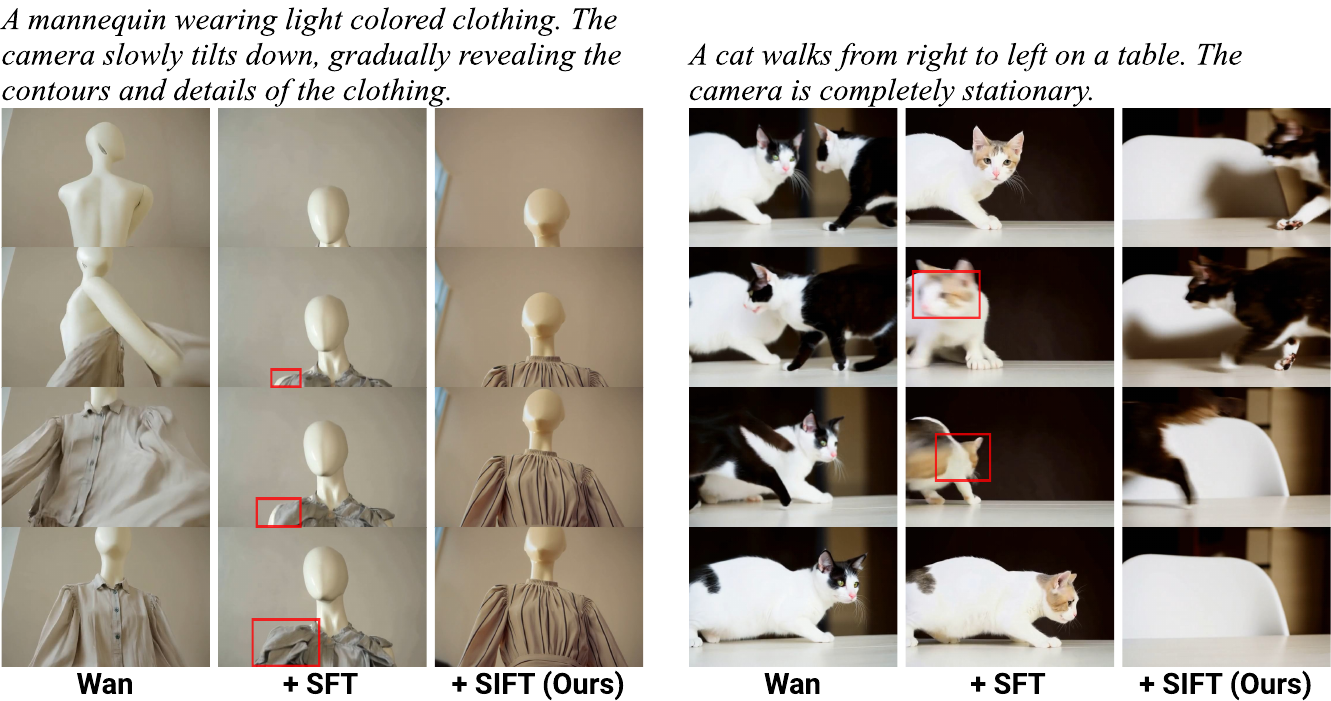}
   \caption{Qualitative comparison on the Wan backbone. SIFT better preserves prompt-specified relative motion and temporal consistency than other baselines.}
   \label{fig: qual_results1}
\end{figure}
\begin{figure}
  \centering
    \includegraphics[width=.6\linewidth]{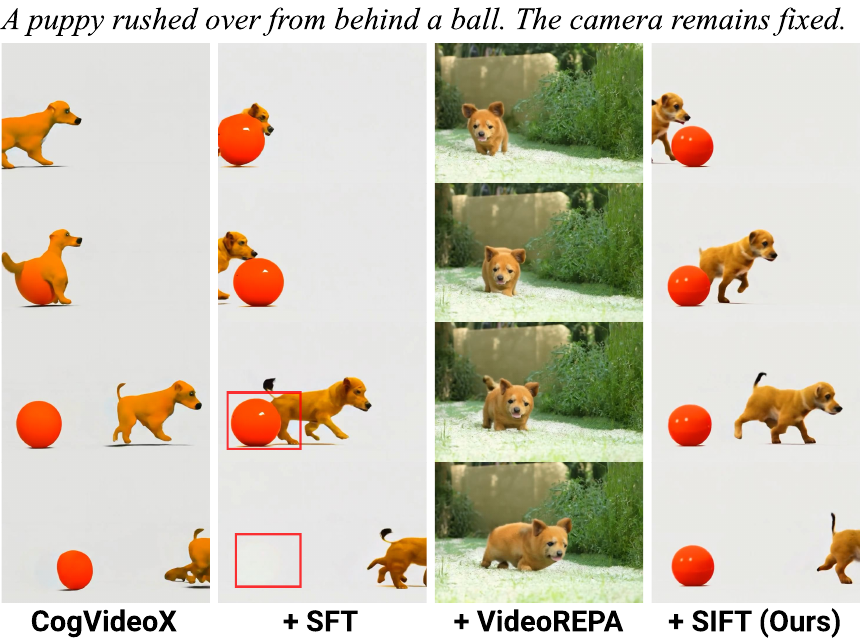}
   \caption{Qualitative comparison on the CogVideoX backbone. SIFT produces more coherent object interactions and more physically plausible relative motion than others.}
   \label{fig: qual_results2}
\end{figure}
\noindent\textbf{Evaluation Metrics.}
As the first work to investigate motion entanglement in videos, we design evaluation metrics based on VideoPhy~\cite{bansal2024videophy} and PhysCtrl~\cite{wang2025physctrl} with necessary adaptations.
We employ both a vision-language model (VLM) and human evaluations to assess each video on 1-5 scale across two dimensions:
(1) \textit{Semantic Adherence (SA)}: measures how well the video content aligns with the text prompt, particularly whether the generated motions match the prompt; (2) \textit{Physical Commonsense (PC)}: measures whether the observed motion follows the physics laws in the real world.
Additionally, we run a pairwise \textit{human preference study}: for each prompt, we present two videos (e.g., our method versus a baseline) and ask evaluators which video better satisfies SA / PC. 

\noindent\textbf{Implementation Details.}
All experiments are conducted on 8 NVIDIA H100 GPUs. We adopt two pre-trained video diffusion models as backbones: Wan2.1-T2V-1.3B~\cite{wan2025wan} and CogVideoX~\cite{yang2024cogvideox}. Both models use an identical learning rate of $5e^{-6}$ and are optimized for 1,000 steps with a batch size of 1. For evaluation, we use InternVideo2.5~\cite{wang2025internvideo2.5} as the VLM evaluator. During the self-imagination phase, video generation proceeds through timesteps $t = 1000, 980, 960$. The Progressive Hard Case Replay warm-up period $T_\text{warmup}$ is set to 500 iterations, with hard case replay performed every 10 batches. The motion-aware discriminative loss weight is set to 0.01. Alternating supervision between motion classifiers is applied every 25 batches.

\subsection{Comparison with Existing Baselines}
\noindent\textbf{Baselines.}
We conduct extensive comparative experiments on two open source video diffusion models: Wan2.1~\cite{wan2025wan} and CogVideoX~\cite{yang2024cogvideox}. The compared methods include: (1) \textit{Original}: the base pre-trained models without any modification; (2) \textit{SFT}: supervised fine-tuning on our human-collected dataset of 4,000 motion-decoupled videos; (3) \textit{VideoREPA~\cite{zhang2025videorepa}}: a method that applies REPA~\cite{yu2024representationrepa} to video diffusion model, distilling general physical understanding from video foundation models by aligning token-level relations; and (4) our proposed \textit{SIFT}. 




\noindent\textbf{Quantitative Evaluation.}
As summarized in \cref{tab: main_results}, both VLM-based and human evaluations across two backbone models demonstrate that our method achieves the best performance in both Semantic Adherence (SA) and Physical Commonsense (PC). 
The SFT baseline brings only limited gains and even underperforms the original model in some human evaluations (e.g., Wan + SFT obtains a PC score of 2.43, versus 2.75 for vanilla Wan).
This supports our claim that conventional SFT helps the model imitate motion distributions present in the data but fails to cultivate genuine physical reasoning. The model tends to memorize statistical correlations of training data but cannot generalize to novel scenarios, often generating kinematically inconsistent motions that human raters can easily spot.
VideoREPA~\cite{zhang2025videorepa} enhances temporal stability by distilling general physical knowledge via feature alignment but lacks explicit supervision for motion disentanglement. As a result, improvements in PC metrics are modest, with smoother motion visuals yet frequent errors in relative dynamics (e.g., camera versus object motion).
In contrast, our method demonstrates significant improvement across all metrics. By breaking the reconstruction shortcut and learning from self-imagined videos, SIFT compels the model to infer motion directly from semantic prompts, rather than replicating residual motion cues.
The motion-aware discriminative supervision and progressive hard-case replay further guide the model to correct mistakes and stabilize learning, leading to robust disentanglement of independent motion sources.
Notably, human preference studies in \cref{fig: preference} confirm that SIFT is strongly favored over all baselines, showing that our method enhances both the realism and controllability of generated motion without sacrificing semantic fidelity.

\noindent\textbf{Qualitative Evaluation.}
The qualitative results between our method and baselines can be found in \cref{fig: qual_results1,fig: qual_results2}, and they further illustrate the advantages of SIFT.
For instance, in the puppy-and-ball scene, SIFT generates natural relative motion and consistent object presence across frames. However, the original CogVideoX exhibits unrealistic interactions where the puppy partially merges with the ball; CogVideoX + SFT produces temporal discontinuities with the ball abruptly disappearing; and CogVideoX + VideoREPA fails to generate the ball altogether. These artifacts demonstrate the limitations of existing approaches in modeling kinematically correct and coherent motion dynamics.

\begin{wraptable}{r}{0.4\linewidth} 
    \vspace{-7mm}
    \caption{Comparison with camera-controlled methods.}
    \centering
    \small
    \scalebox{1}{ 
    \begin{tabular}{llccc}
    \toprule
Method 
& SA $\uparrow$
& PC $\uparrow$ \\
\midrule
CameraCtrl  &3.07 & 3.03 \\
Wan-Move &3.73&3.70\\
SIFT &\textbf{4.10} & \textbf{4.30} \\
\bottomrule
\end{tabular}}
\label{table:comercial}
\end{wraptable}
\noindent\textbf{Camera-controlled Methods.} Camera-controlled methods~\cite{he2024cameractrl,chu2025wan-move} primarily target precise controllability through dense external conditions, such as manually designed trajectories and reference first frames, whereas SIFT aims to improve the backbone model’s intrinsic motion prior under the standard text-to-video setting. 
For completeness, we report a separate comparison on a camera-only subset in ~\cref{table:comercial}, where we align the inputs as closely as possible: camera poses are specified using a small set of fixed directional trajectories, and required reference first frames are generated by Nano Banana~\cite{google_nanobanana}.

\begin{wraptable}{r}{0.5\linewidth}
    \vspace{-4mm}
    \caption{Generalization to more complex motion settings.}
    \label{tab:generalization}
    \centering
    \scriptsize
    \resizebox{0.8\linewidth}{!}{
    \begin{tabular}{llcc}
    \toprule
    Setting & Method & SA $\uparrow$ & PC $\uparrow$ \\
    \midrule
    \multirow{2}{*}{Multi-object}
        & Wan & 3.88 & 3.92 \\
        & + SIFT & \textbf{4.50} & \textbf{4.18} \\
    \midrule
    \multirow{2}{*}{Articulated}
        & Wan & 3.76 & 4.04 \\
        & + SIFT & \textbf{4.02} & \textbf{4.36} \\
    \midrule
    \multirow{2}{*}{Long horizon}
        & Wan & 3.50 & 3.82 \\
        & + SIFT & \textbf{3.88} & \textbf{4.31} \\
    \bottomrule
    \end{tabular}}
\end{wraptable}
\noindent\textbf{Generalization Beyond Basic Motion Disentanglement.}
Beyond the camera-only and object-only settings in our main benchmark, we further evaluate SIFT on three more challenging motion scenarios: multi-object motion, articulated motion, and long-horizon motion. Each setting contains 50 prompts. For the long-horizon setting, each prompt involves at least two sequential motion phases, and the generated videos contain twice as many frames as those in the standard benchmark. As shown in Tab.~\ref{tab:generalization}, SIFT consistently improves both semantic adherence and physical commonsense across all three settings. These results suggest that SIFT improves the model's understanding of motion, enabling it to achieve better performance in more complex and general scenarios.


\begin{table}[t]
    \centering
    \small
    \caption{Ablation study of SIFT components on the Wan using VLM evaluation. SA: semantic adherence; PC: physical commonsense. For a fair and accurate comparison, videos generated from the same prompt by all methods are jointly provided to the VLM for scoring. \textbf{So \textbf{s}cores are relative within each group and should not be directly comparable to those in \cref{tab: main_results}.}}
    \begin{tabular}{lcccc} 
        \toprule
        \multirow{2}{*}{\textbf{Model Variant}} & \multicolumn{2}{c}{\textbf{Camera-only}} & \multicolumn{2}{c}{\textbf{Object-only}} \\
        \cmidrule(lr){2-3} \cmidrule(lr){4-5}
         & \textbf{SA($\uparrow$)} & \textbf{PC($\uparrow$)} & \textbf{SA($\uparrow$)} & \textbf{PC($\uparrow$)} \\
        \midrule
        
        \rowcolor{gray!10}\multicolumn{5}{l}{\textbf{(a) Self-Imagination Generation}} \\
        w/ Reconstruction Shortcut      &3.05&3.55&3.45&3.65\\
        
        \addlinespace[4pt]
        \rowcolor{gray!10}\multicolumn{5}{l}{\textbf{(b) Alternating Supervision}} \\
        w/ Single Classifier (R3D)         &3.65&3.55&3.20&3.60\\
        w/ Single Classifier (SlowFast)    &3.35&3.25&3.50&3.85\\
        
        \addlinespace[4pt]
        \rowcolor{gray!10}\multicolumn{5}{l}{\textbf{(c) Progressive Hard Case Replay}} \\
        w/o Progressive Hard Case Replay          &3.00&3.45&3.35&3.70\\
        
        \midrule
        \textbf{Ours (Full SIFT)}       &\textbf{3.85}&\textbf{4.15}&\textbf{3.50}&\textbf{4.00}\\
        \bottomrule
    \end{tabular}
    \label{tab: ablation}
\end{table}

\subsection{Ablation Study}

To better understand the contribution of each component of our SIFT, we conducted a
comprehensive ablation study, as shown in \cref{tab: ablation}. All ablations are evaluated on a randomly sampled set of 40 prompts from the test benchmark.

\noindent\textbf{Self-Imagination Generation.}
We ablate the self-imagination design by initializing generation from noisy real videos instead of pure noise, while keeping all other components unchanged.
As shown in \cref{tab: ablation}, this variant performs worse in both semantic adherence and physical plausibility.
When conditioned on noised real videos, the model exploits residual motion cues, such as coarse object trajectories and temporal coherence, to guide reconstruction, bypassing the need to infer motion from text. Moreover, because real videos are inherently kinematically plausible, the model tends to copy motion patterns present in the input rather than learning to infer kinematically plausible motion dynamics. As a result, it fails to develop a robust understanding of motion semantics from language, leading to poorer generalization and reduced motion disentanglement.

\begin{figure}[t]
  \centering
    \includegraphics[width=.6\linewidth]{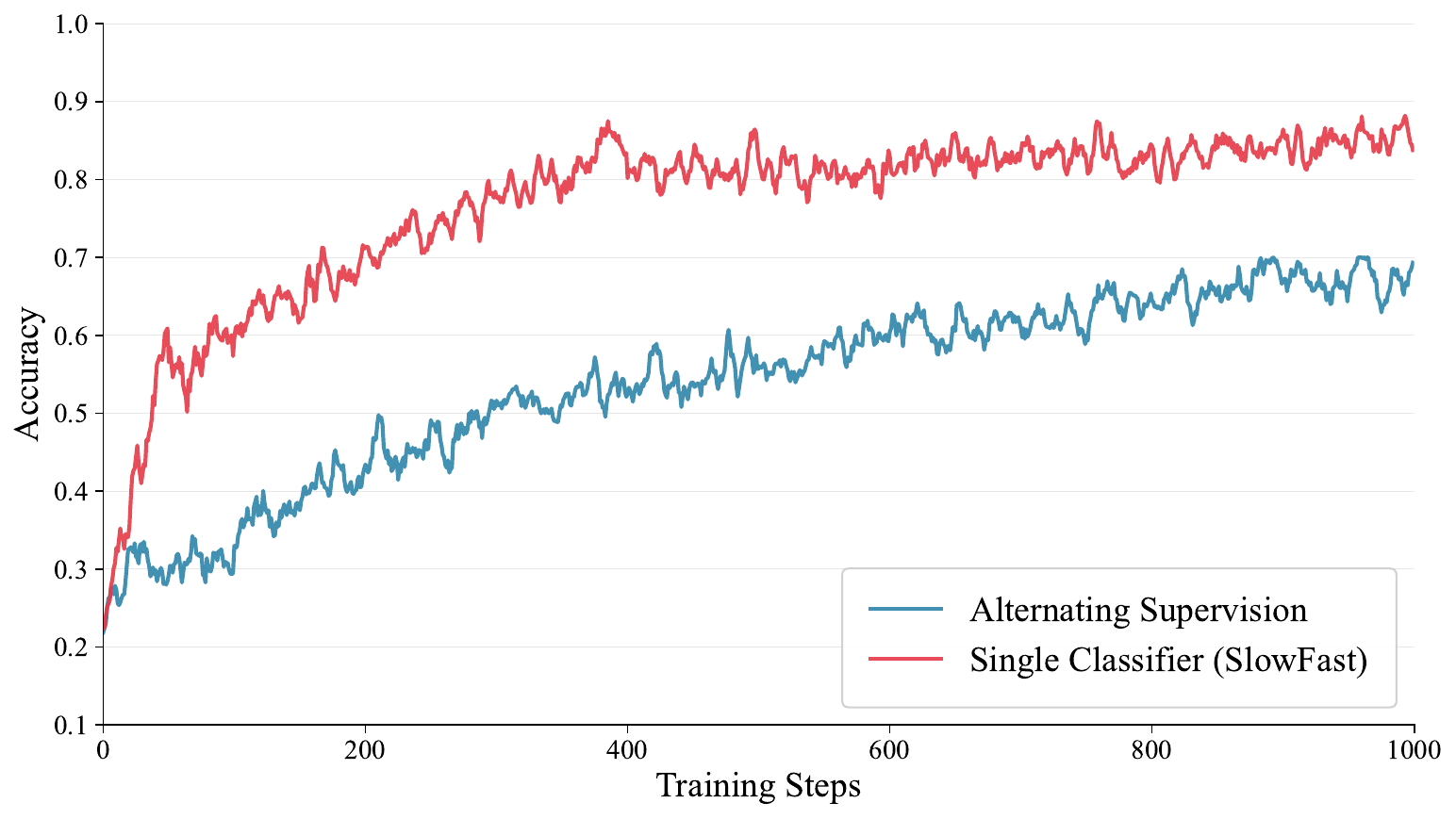}
   \caption{Accuracy curves of motion classifiers during training of video generative model.}
   \label{fig: ablation_acc}
\end{figure}

\noindent\textbf{Alternating Supervision.}
We evaluate our alternating strategy by comparing it against using only a single motion classifier supervision (R3D or SlowFast). As shown in \cref{tab: ablation}, both single-classifier variants perform worse than our full approach. 
As illustrated in \cref{fig: ablation_acc}, training with a single classifier causes its accuracy to saturate quickly, indicating overfitting to its own inductive bias.
This can mislead the generator to reproduce artifacts favored by that classifier rather than generating physically plausible motion.
In contrast, alternating between heterogeneous classifiers prevents overfitting to a fixed inductive prior and encourages the generator to satisfy complementary motion criteria, resulting in more stable and kinematically coherent motion generation.

\noindent\textbf{Progressive Hard Case Replay.}
We ablate this component by removing both the hard case buffer and the progressive scheduling, i.e., exposing the model to hard cases from the beginning of training. As shown in \cref{tab: ablation}, this variant leads to clear degradation in both physical plausibility and semantic adherence. The performance drop occurs because introducing challenging cases too early injects noisy gradients when the model's parameters are not yet sufficiently prepared. This disrupts motion learning and even harms overall visual quality, as evidenced by the more significant decline in semantic adherence (SA).
In contrast, our progressive strategy allows the model to first master basic motion patterns before tackling more challenging cases, leading to more stable convergence and better final performance. 
Moreover, the hard case replay improves training efficiency, enabling the model to achieve greater improvement within the same number of optimization steps.



\section{Conclusion}
\label{sec:conclusion}

In this work, we identify a key kinematic limitation in current video diffusion models, which we term ``Motion Entanglement''. We show that this issue arises from data-induced spurious correlations as well as the reconstruction shortcut and objective biases inherent in their training paradigm. These factors prevent models from learning physically grounded motion. To address this, we propose Self-Imagination Fine-Tuning (SIFT), a paradigm shift from reconstruction-based training to imagination-driven learning that initializes generation from random noise. Combined with motion-aware discriminative supervision and a progressive hard-case replay strategy, SIFT enables the model to infer, evaluate, and correct motion dynamics without relying on scarce motion-decoupled data. Extensive experiments on both Wan and CogVideoX demonstrate that SIFT significantly improves physical plausibility and motion disentanglement while preserving semantic fidelity, paving the way toward more realistic and controllable text-to-video generation.

\noindent\textbf{Limitations and Future Work.} Although the four-way motion taxonomy provides effective supervision for disentangling camera and object motion, it remains coarse. In particular, it does not explicitly model motion direction, magnitude, individual trajectories, or temporal transitions between multiple motion states. This limitation becomes more evident in involving multiple moving entities and complex motion composition scenarios, where a single categorical label may not fully describe the underlying motion structure. In addition, the quality of the feedback depends on the robustness of the motion classifiers, which can be uncertain for ambiguous or severely corrupted self-generated videos. A promising direction is to introduce richer factorized motion representations, such as object-level trajectories, optical flow, or motion fields, and to design more fine-grained feedback for complex interactions and longer temporal horizons.



%
%
\bibliographystyle{splncs04}
\bibliography{main}

\appendix

\clearpage
\setcounter{page}{1}
\setcounter{table}{0}
\setcounter{figure}{0}


\renewcommand{\thetable}{S\arabic{table}}
\renewcommand{\thefigure}{S\arabic{figure}}

\section*{Overview} 
This supplementary material provides additional implementation details, including the LLMs used for prompt generation and evaluation, together with a full description of our human study setup and additional experimental results that were not included in the main paper.
\textbf{We strongly encourage readers to watch the video}, as the key phenomena and improvements are most clearly observable in dynamic motion.
 
\section{LLM/VLM Implementation Details}
This section provides the detailed settings, prompts, and procedures for all LLM/VLM components used in our work, including (1) VLM-based evaluation, (2) training prompt generation, and (3) testing prompt generation.

\noindent \textbf{InternVideo2.5 Evaluation.}
We adopt InternVideo2.5~\cite{wang2025internvideo2.5} as an automatic evaluator to assess Semantic Adherence (SA) and Physical Commonsense (PC) of videos generated by all compared methods.
To ensure fair comparison, for each prompt, we collect videos from all methods, shuffle their order, and jointly present them to the VLM along with the text prompt. We then prompt InternVideo2.5 with the following prompt to use it for evaluation:

\begin{tcolorbox}[
  colback=gray!10,    
  colframe=gray!70,   
  boxrule=0.5pt,      
  title={Video Evaluation},
  fonttitle=\bfseries,
]
You are comparing several versions of the same video scene, each generated by different methods.
The order of videos corresponds to: \textit{\{subdirs\}}.

For each version, rate the video along the following two dimensions, each using a 5-point Likert scale
(1 = poor, 5 = excellent). Provide only numerical scores and no explanations.

1. Semantic Adherence \\
Rate how well the video matches the prompt:
\begin{itemize}[leftmargin=*]
    \item Objects, actions, and events correspond to the prompt (penalize missing or extra elements).
    \item Scene layout, background, and interactions align with the prompt.
\end{itemize}

2. Physical Plausibility \\
Rate how realistic the motions are:
\begin{itemize}[leftmargin=*]
    \item Motions are continuous, stable, and physically possible in the real world.
    \item Objects move only with plausible causes (no drifting or sudden changes without force).
\end{itemize}

\end{tcolorbox}

\noindent \textbf{GPT-4o Training prompts Generation.}
We use GPT-4o~\cite{hurst2024gpt4o} to generate 10,000 training prompts covering diverse camera-only/object-only motion combinations.
We use the following instruction:

\begin{tcolorbox}[
  colback=gray!10,
  colframe=gray!70,
  boxrule=0.5pt,
  title={Camera-Only Motion Prompt Generation},
  fonttitle=\bfseries,
]
Generate diverse and unique prompts for video generation where only the camera moves. Each prompt must consist of two sentences: 
\begin{enumerate}[leftmargin=*]
    \item A description of a scene with completely stationary subjects (no movement). 
    \item A description of the camera action, starting with ``The camera ...''.
\end{enumerate}

Requirements:
\begin{itemize}[leftmargin=*]
    \item Subjects should be diverse: vehicles, animals, people, objects, furniture, sports equipment, artworks, etc.
    \item Camera movements should vary and be selected only from: pan, tilt, zoom, truck, pedestal, dolly, arc, orbit, rotate.
    \item Use simple, natural English. Avoid rare or overly technical film terms.
    \item Avoid repeating the same subject or camera movement too often.
\end{itemize}
\end{tcolorbox}

\begin{tcolorbox}[
  colback=gray!10,
  colframe=gray!70,
  boxrule=0.5pt,
  title={Object-Only Motion Prompt Generation},
  fonttitle=\bfseries,
]
Generate diverse and unique prompts for video generation where only the subjects move. Each prompt should include:
\begin{enumerate}[leftmargin=*]
    \item A description of a scene with moving subjects.
    \item A description of a completely static camera starting with ``The camera''.
\end{enumerate}

Requirements:
\begin{itemize}[leftmargin=*]
    \item Subjects should be highly diverse: animals, people, vehicles, objects, etc.
    \item Subject movements can be active (run, move, fly, drive, rotate, etc.) or passive (fall, flow, swing, being thrown, drift, etc.).
    \item Ensure no two prompts are similar in subject or movement type.
\end{itemize}
\end{tcolorbox}

\noindent \textbf{Gemini-Pro-Vision-1.5 Testing prompt Generation.}
To avoid train–test contamination and to ensure fair evaluation, we use a different modelGemini Pro Vision 1.5~\cite{team2024gemini}, to generate testing prompts. We request the LLM to produce prompts using the same instructions as GPT-4o.
\begin{figure*}[t]
  \centering
  \includegraphics[width=\linewidth]{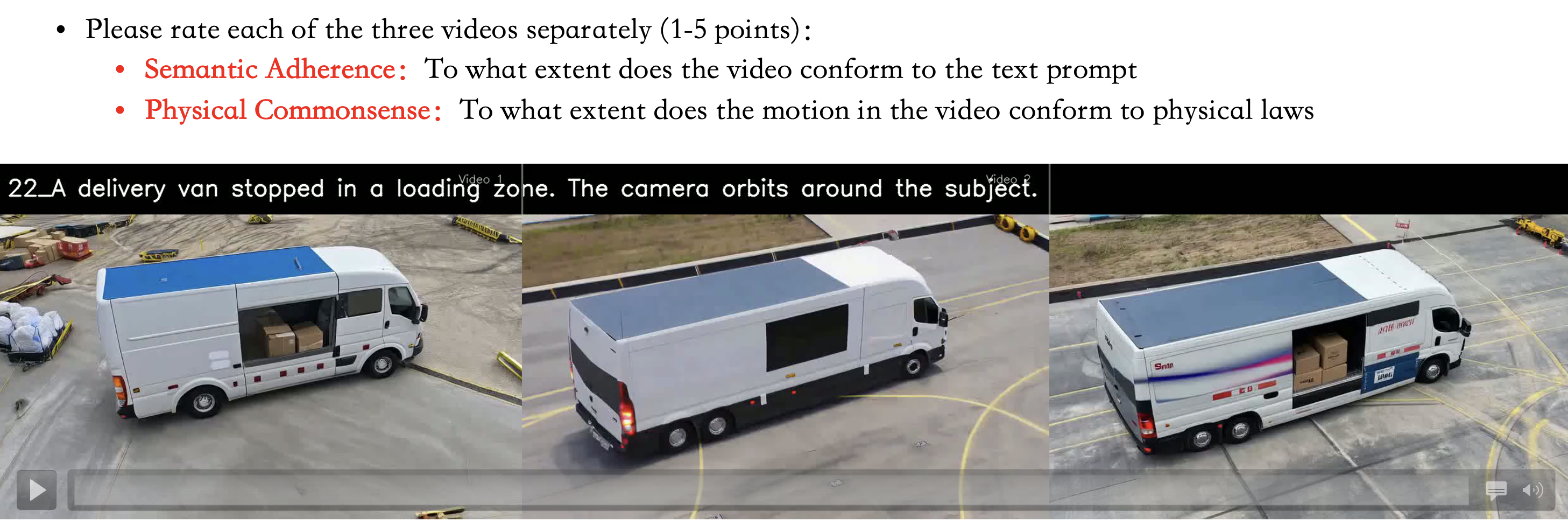}
   \caption{Illustration of the user interface used for evaluating video generations.}
   \label{fig: sup-user_interface}
\end{figure*}
\section{More Results}
For clearer visualization and deeper qualitative understanding, we provide extensive video results in the supplementary materials.

\section{Human Evaluation Details}
We conduct two complementary forms of human evaluation to both Semantic Adherence (SA) and Physical Commonsense (PC):
(1) Likert-scale scoring, and (2) pairwise preference judgments.

In the Likert-scale evaluation, participants are presented with a text prompt and all corresponding videos from compared methods (in randomized order), and are asked to rate each video individually on a 1–5 scale for both SA and PC. In the pairwise evaluation, participants are shown two videos (ours vs. a baseline) alongside the prompt and must select which one better satisfies SA and PC, respectively.

The evaluation interface is illustrated in Figure~\ref{fig: sup-user_interface}. We collected 20 complete responses, with each participant evaluating 16 video groups (i.e., 20 participants $\times$ 16 comparisons). For the Likert-scale evaluation, we report the average scores across all participants. For the pairwise evaluation, we report the average winning percentage of our method over each baseline.

\section{Failure Cases}
\begin{wrapfigure}{r}{0.6\textwidth} 
  \centering
  \includegraphics[width=0.6\textwidth]{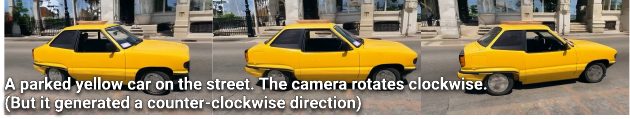}
   \caption{An example of failure cases.}
   \label{fig:failure}
\end{wrapfigure}
While SIFT substantially mitigates the physically implausible coupling between camera and object motions, we observe that the camera may move in the wrong direction. For instance, as shown in \cref{fig:failure}, when a prompt specifies a ``clockwise" camera trajectory, the camera may execute a counter-clockwise movement instead. This failure appears to stem from limitations in the backbone model’s directional semantic grounding.
\section{More Results}
For clearer visualization and deeper qualitative understanding, we provide extensive video results in the supplementary materials.

\end{document}